\documentclass{article}

\PassOptionsToPackage{numbers,sort&compress,square}{natbib}
 \usepackage[final]{neurips_2025}

\usepackage[utf8]{inputenc} % allow utf-8 input
\usepackage[T1]{fontenc}    % use 8-bit T1 fonts
\usepackage{hyperref}       % hyperlinks
\usepackage{url}            % simple URL typesetting
\usepackage{booktabs}       % professional-quality tables
\usepackage{amsfonts}       % blackboard math symbols
\usepackage{nicefrac}       % compact symbols for 1/2, etc.
\usepackage{microtype}      % microtypography
\usepackage{xcolor}         % colors
\usepackage{graphicx}
\usepackage{array}
\usepackage{colortbl} 
\usepackage{adjustbox}
\usepackage{listings}
\usepackage{caption}

\usepackage{float} 
\usepackage{placeins}  
\usepackage{dblfloatfix} 
\usepackage{amsmath}
\usepackage{amssymb}
\usepackage{mathtools}
\usepackage{amsthm}
\usepackage{subcaption}
\usepackage{enumitem}
\usepackage{multirow}
\usepackage[dvipsnames]{xcolor}

\title{ARIAL: An Agentic Framework for Document VQA with Precise Answer Localization}

\author{
Ahmad Mohammadshirazi\\
Ohio State University\\
Flairsoft\\
Columbus, Ohio, US\\
{\tt\small mohammadshirazi.2@osu.edu}
\and
Pinaki Prasad Guha Neogi\\
Ohio State University\\
Columbus, Ohio, US\\
{\tt\small guhaneogi.2@osu.edu}
\and
Dheeraj Kulshrestha\\
Flairsoft\\
Columbus, Ohio, US\\
{\tt\small dheeraj@flairsoft.net}
\and
Rajiv Ramnath\\
Ohio State University\\
Columbus, Ohio, US\\
{\tt\small ramnath.6@osu.edu}
}

\begin{document}

\maketitle

\begin{abstract}
Document Visual Question Answering (VQA) requires models to not only extract accurate textual answers but also precisely localize them within document images—a capability critical for interpretability in high-stakes applications. However, existing systems achieve strong textual accuracy while producing unreliable spatial grounding, or sacrifice performance for interpretability. We present ARIAL (\textbf{A}gentic \textbf{R}easoning for \textbf{I}nterpretable \textbf{A}nswer \textbf{L}ocalization), a modular framework that orchestrates specialized tools through an LLM-based planning agent to achieve both precise answer extraction and reliable spatial grounding. ARIAL decomposes Document VQA into structured subtasks: OCR-based text extraction with TrOCR, retrieval-augmented context selection using semantic search, answer generation via fine-tuned Gemma 3-27B, and explicit bounding-box localization through text-to-region alignment. This modular architecture produces transparent reasoning traces, enabling tool-level auditability and independent component optimization. We evaluate ARIAL on four benchmarks—DocVQA, FUNSD, CORD, and SROIE—using both textual accuracy (ANLS) and spatial precision (mAP@IoU 0.50:0.95). ARIAL achieves SoTA results across all datasets: 88.7 ANLS and 50.1 mAP on DocVQA, 90.0 ANLS and 50.3 mAP on FUNSD, 85.5 ANLS and 60.2 mAP on CORD, and 93.1 ANLS on SROIE, surpassing the previous best method (DLaVA) by +2.8 ANLS and +3.9 mAP on DocVQA. Our work demonstrates how agentic orchestration of specialized tools can simultaneously improve performance and interpretability, providing a pathway toward trustworthy, explainable document AI systems. Code is available at: \url{https://github.com/ahmad-shirazi/ARIAL}
\end{abstract}     
\section{Introduction}
\label{sec:intro}

\begin{figure*}[h]
    \centering
    \includegraphics[width=1\textwidth]{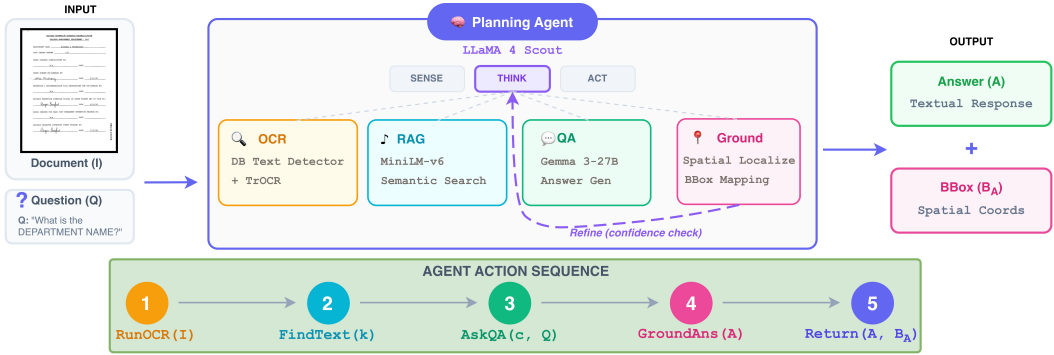}
\caption{Overview of the ARIAL agentic workflow for Document VQA. The system consists of three modular stages: (1) Input Processing, where an OCR module extracts text segments and bounding boxes from a document image; (2) Agentic Reasoning Pipeline, where the planner agent coordinates task execution—retrieving relevant text, invoking QA or computation, and triggering spatial grounding; and (3) Output Generation, where the final answer and its bounding box are produced. The reasoning loop enables iterative refinement based on confidence, supporting flexible and context-aware decision-making.}
    \label{fig:pipeline}
\end{figure*}

Document Visual Question Answering (VQA) requires reasoning over both textual content and visual layout in scanned or digitally rendered documents. Models must not only read and understand diverse formats—forms, receipts, reports—but also locate where answers appear within the document structure.

While recent models such as LayoutLMv3~\cite{huang2022layoutlmv3}, LayoutLLM~\cite{luo2024layoutllm}, and DocLayLLM~\cite{liao2025doclayllm} have improved textual accuracy by combining language with layout features, they often treat localization as a secondary task. Consequently, they may generate plausible answers without clearly identifying their source in the document, making verification difficult. Standard metrics like ANLS~\cite{yujian2007normalized} capture string similarity but fail to reflect spatial correctness, prompting a shift towards combined evaluations that include IoU for grounding precision.

DLaVA~\cite{mohammadshirazi2025dlava} introduced answer localization by integrating bounding-box prediction within a large multimodal transformer. However, its monolithic design can be computationally intensive and may struggle with fine-grained details in dense or handwritten layouts.

We propose \textbf{ARIAL} (Agentic Reasoning for Interpretable Answer Localization), a modular document VQA framework built around an agentic planning model. Rather than using a single large model, ARIAL delegates subtasks—OCR, layout analysis, retrieval, reasoning, and grounding—to specialized modules orchestrated by a central agent. This agent, implemented with LLaMA 4 Scout~\cite{llama4scout}, dynamically selects tools and composes multi-step reasoning chains, enabling accurate and interpretable answers with precise spatial grounding. Our key contributions are:

\begin{enumerate}
    \item \textbf{Agentic Document QA}: We introduce an agent-based document VQA system that decomposes queries into tool calls for OCR, retrieval, and grounding. The modular design enables tool reuse, error tracing, and flexible adaptation across document types.
    
    \item \textbf{Precise Answer Localization}: ARIAL produces both answer text and corresponding bounding boxes by aligning answers to OCR-detected spans and contextual cues, ensuring visual traceability.
    
    \item \textbf{Retrieval-Augmented Reasoning}: ARIAL incorporates retrieval-augmented generation~\cite{lewis2020retrieval} to focus on relevant text segments, enhancing both reasoning accuracy and efficiency for long or noisy documents.
    
    \item \textbf{SoTA Results}: On four benchmarks—DocVQA~\cite{mathew2021docvqa}, FUNSD~\cite{jaume2019funsd}, CORD~\cite{park2019cord}, and SROIE~\cite{huang2019icdar2019}—ARIAL achieves new best results in both ANLS and mAP@IoU, reaching 88.7 ANLS and 50.1 mAP on DocVQA.
\end{enumerate}

ARIAL demonstrates how LLMs can be effectively constrained through modular tool orchestration, where each answer is locked to specific pixel coordinates and traceable through interpretable reasoning chains. This addresses fundamental challenges in developing trustworthy, location-aware AI systems for document understanding.

The remainder of this paper is organized as follows: Section~\ref{sec:related} reviews related work in document VQA and agentic AI. Section~\ref{sec:methodology} details ARIAL's architecture and modules. Section~\ref{sec:4} outlines datasets and evaluation protocols. Results and analysis appear in Section~\ref{sec:results}, followed by discussion in Section~\ref{sec:discussion} and conclusions in Section~\ref{sec:conclusion}.

\begin{figure*}[h]
  \centering
  \begin{subfigure}[b]{0.30\textwidth}
    \centering
    \includegraphics[width=\linewidth, trim={0cm 4.5cm 0cm 0cm}, clip]{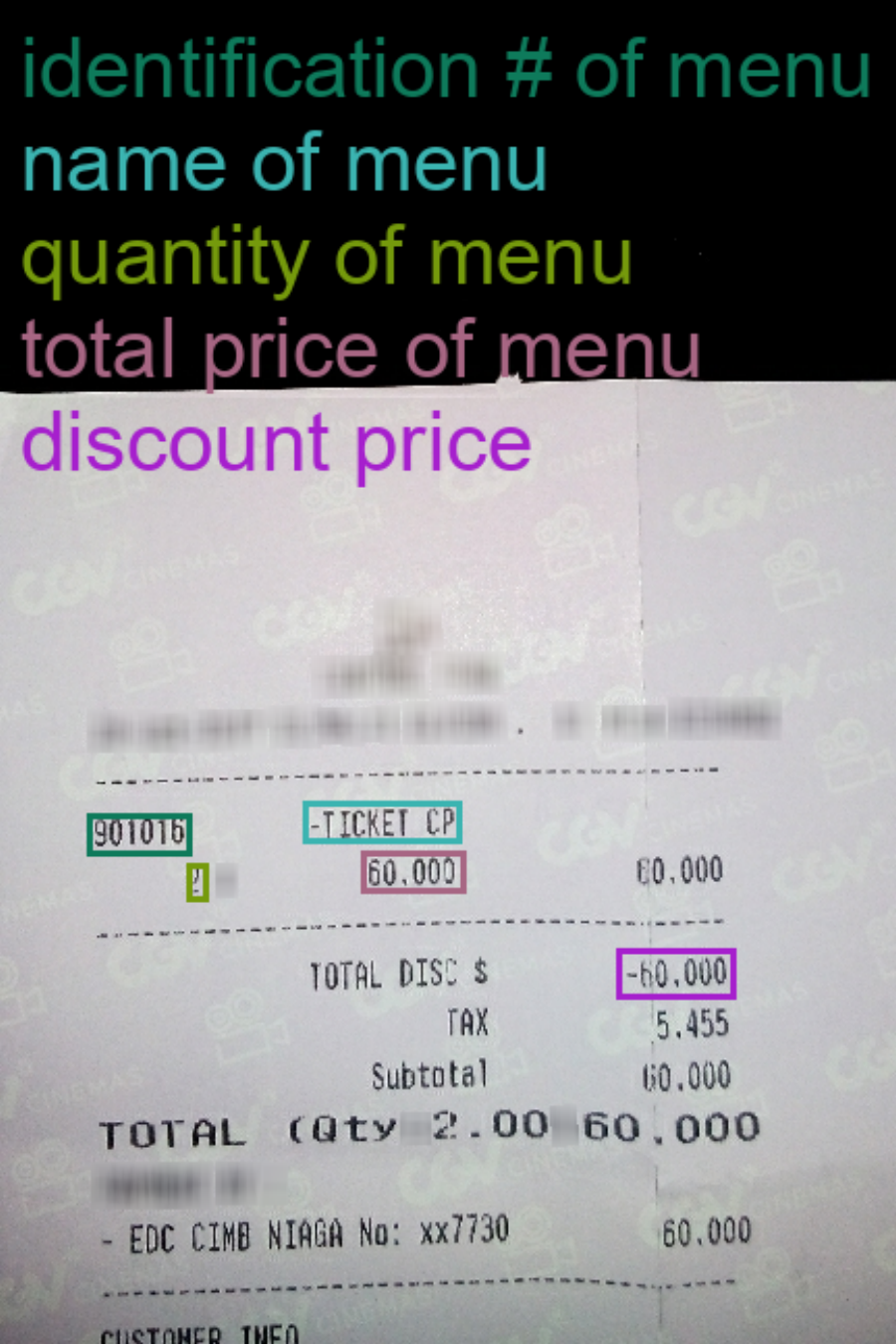}
  \end{subfigure}\hfill
  \begin{subfigure}[b]{0.33\textwidth}
    \centering
    \includegraphics[width=\linewidth, trim={0cm 4.5cm 0cm 0cm}, clip]{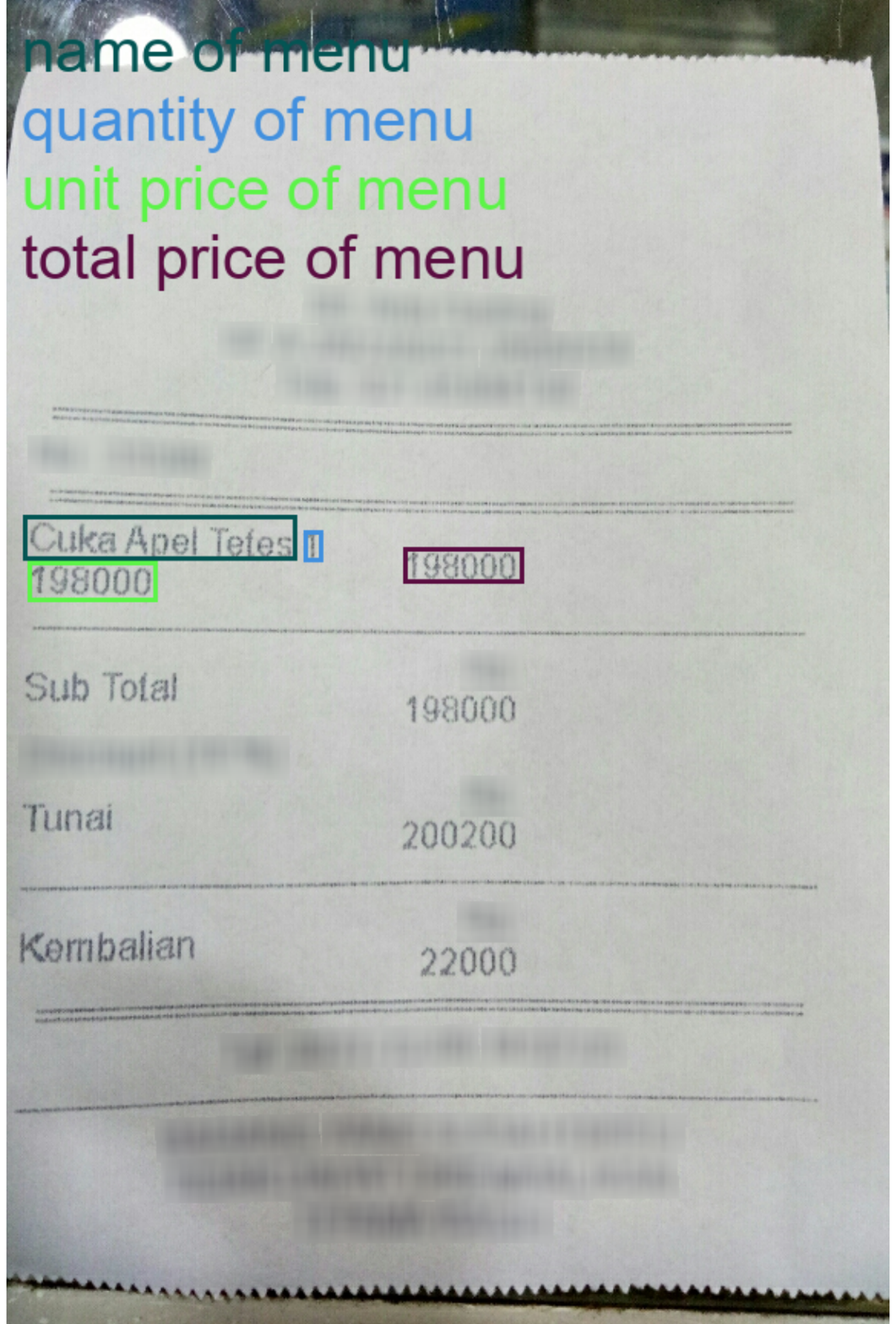}
  \end{subfigure}\hfill
  \begin{subfigure}[b]{0.33\textwidth}
    \centering
    \includegraphics[width=\linewidth, trim={0cm 4.5cm 0cm 0cm}, clip]{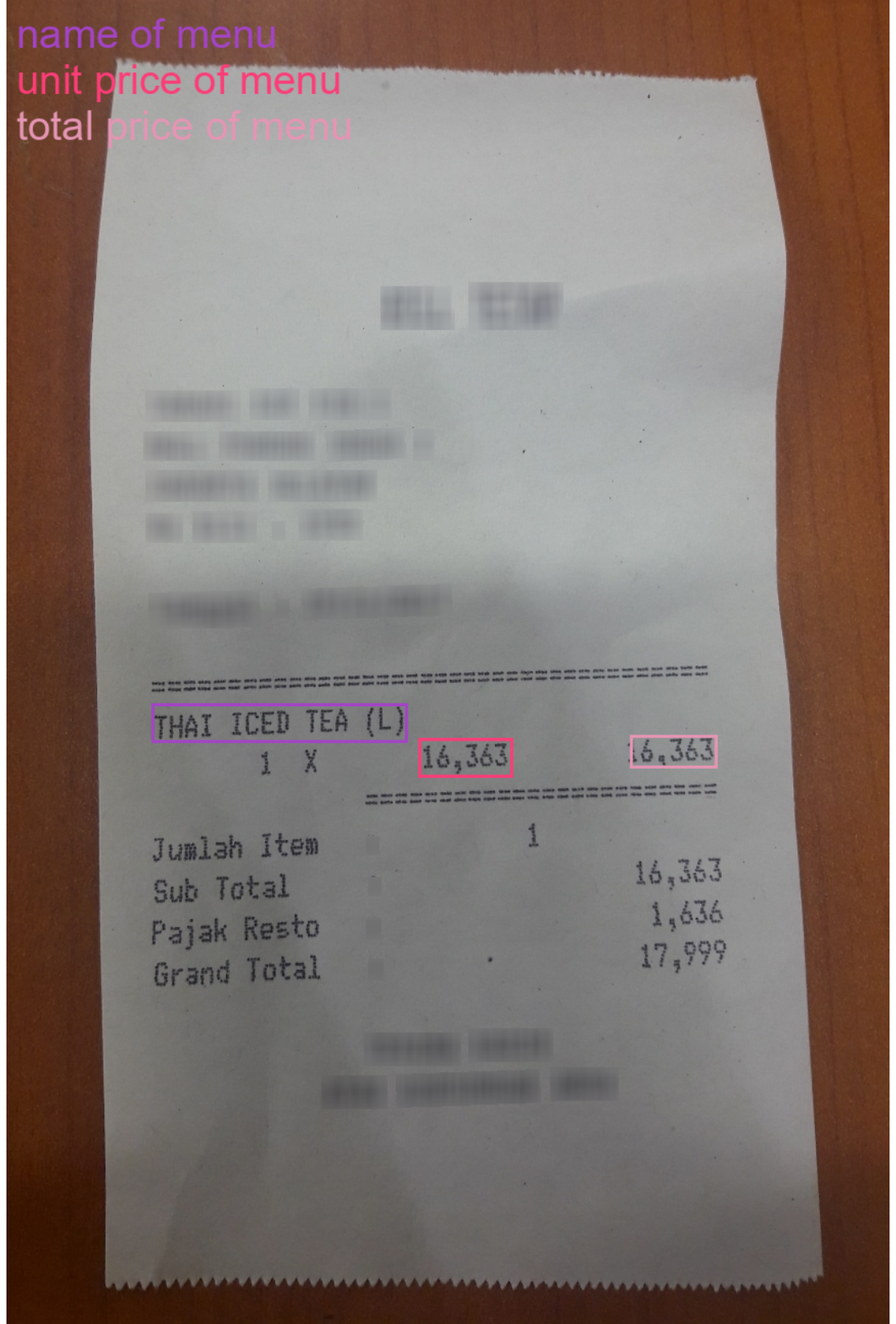}
  \end{subfigure}
  \caption{Illustrative examples of visual information extraction on receipt images from the CORD dataset~\cite{park2019cord}. Each colored annotation corresponds to its extracted answer, highlighted by a matching colored bounding box.}
  \label{fig:sample}
\end{figure*}
\section{Related Work}
\label{sec:related}

\subsection{Document VQA and Layout-Aware Models}

Early document QA systems treated the task as text-only reading comprehension by applying OCR and feeding results into standard NLP models~\cite{mishra2019ocr}. However, such approaches ignored document structure, prompting the development of layout-aware models. LayoutLM~\cite{xu2020layoutlm}, LayoutLMv3~\cite{huang2022layoutlmv3}, DocFormer~\cite{appalaraju2021docformer}, and StrucTexT~\cite{li2021structext} embed both text and spatial coordinates to model document layouts more effectively, achieving strong performance on datasets like DocVQA through unified transformer architectures.

Nevertheless, most models output only answer text and treat localization as auxiliary prediction or post-hoc mapping. Methods like TILT~\cite{powalski2021going} and Donut~\cite{kim2022ocr} explore end-to-end generation—Donut bypasses explicit OCR—but lack transparent mechanisms for spatial grounding. As highlighted by DLaVA~\cite{mohammadshirazi2025dlava}, the inability to visualize answer provenance limits model interpretability and hinders error analysis in high-trust domains.

\subsection{Multimodal LLMs for Documents}

Multimodal large language models (MLLMs) such as GPT-4o~\cite{hurst2024gpt}, Gemini 2.5 Pro~\cite{google_cloud_gemini2.5pro}, and LLaVA 1.5~\cite{liu2024improved} extend VQA capabilities by jointly modeling vision and language. These systems answer questions directly from document images using prompt-based interfaces but often function as black boxes, lacking explicit reasoning steps and failing to highlight the visual basis of their answers. Their reliance on global visual understanding can lead to errors in fine-grained text recognition and spatial disambiguation~\cite{bai2024hallucination}.

Recent domain-specific adaptations like LayoutLLM~\cite{luo2024layoutllm} augment prompts with structured spatial cues to guide model focus. DLaVA~\cite{mohammadshirazi2025dlava} combines detected text with bounding box metadata or constructed text images, enabling prediction of both answer and spatial location. While DLaVA improves interpretability, it relies on a large, end-to-end multimodal backbone. Our method adopts a modular agentic design enabling more transparent and controllable reasoning while retaining compatibility with any OCR or LLM module.

\subsection{Agent-Based and Modular Reasoning}

Agentic frameworks have emerged as powerful alternatives to monolithic models for complex tasks~\cite{hayawi2024ai}. Systems like HuggingGPT~\cite{shen2023hugginggpt} use a central language model to coordinate multiple tools for multi-step reasoning. Multi-agent paradigms have been explored for general VQA~\cite{zhang2025igniting}, where specialized agents handle subtasks like reading, counting, or visual interpretation. HAMMR~\cite{castrejon2024hammr} introduces hierarchical architecture improving reasoning granularity and debuggability.

In the document domain, MDocAgent~\cite{han2025mdocagent} employs multiple agents for long-document QA with roles spanning retrieval, modality-specific analysis, key information extraction, and summarization. This modular approach demonstrated notable performance gains, showing the potential of agentic decomposition.

ARIAL builds upon these foundations by tailoring an agentic framework for document VQA. Unlike generic VQA agents, ARIAL handles document-specific challenges such as dense typography, noisy scans, and form-based structures. Its modularity allows independent component upgrades, facilitating efficient domain adaptation and improving interpretability. ARIAL advances document understanding by combining the reasoning power of MLLMs with the transparency and controllability of agentic pipelines, enabling precise answer localization and robust performance across diverse document types.
\section{Methodology}
\label{sec:methodology}

\subsection{Overview}

\begin{table*}[t]
\centering
\caption{Performance comparison on Document VQA datasets using ANLS (textual accuracy).}
\label{tab:anls_results}
\scriptsize
\resizebox{\textwidth}{!}{
\begin{tabular}{l|l|c|c|c|c}
\toprule
\textbf{Category} & \textbf{Method} 
& \textbf{DocVQA} 
& \textbf{FUNSD} 
& \textbf{CORD} 
& \textbf{SROIE} \\
\midrule

\multirow{2}{*}{Text Only} 
& Llama2-7B-Chat~\cite{touvron2023llama2} 
& 64.99 & 48.20 & 47.70 & 68.97 \\
& Llama3-8B-Instruct~\cite{dubey2024llama3} 
& 51.79 & 68.57 & 52.31 & 61.24 \\
\midrule

Text + BBox 
& LayTextLLM~\cite{lu2024laytextllm} 
& 72.83 & 78.65 & 70.81 & 83.27 \\
\midrule

\multirow{7}{*}{Image Only} 
& gpt-oss-20b~\cite{agarwal2025gpt}
& 79.84 & 77.64 & 77.03 & 80.12 \\
& Llama3.2-11B~\cite{dubey2024llama3}
& 78.40 & 65.02 & 42.96 & 61.42 \\
& Pixtral-12B~\cite{agrawal2024pixtral}
& 80.71 & 78.26 & 79.08 & 82.24 \\
& LLaVA-NeXT-13B~\cite{liu2023llava}
& 51.01 & 19.71 & 33.50 & 13.41 \\
& LLaVA-OneVision-7B~\cite{li2024llavaonevision}
& 47.59 & 22.82 & 32.43 & 12.10 \\
& Qwen2.5-VL-7B~\cite{bai2025qwen25vl}
& 68.54 & 58.44 & 39.01 & 56.37 \\
& InternVL2-8B~\cite{chen2024internvl2}
& 71.26 & 57.58 & 55.88 & 81.55 \\
\midrule

BBox + Image
& DLaVA (Pixtral-12B)~\cite{mohammadshirazi2025dlava} 
& 85.9 & 87.6 & 84.4 & 91.4 \\

\midrule
\multirow{5}{*}{Text + BBox + Image} 
& LayoutLLM-7B CoT~\cite{luo2024layoutllm}
& 74.25 & 78.65 & 62.21 & 70.97 \\
& LayoutLLM-7B CoT (Vicuna)~\cite{luo2024layoutllm}
& 74.27 & 79.98 & 63.10 & 72.12 \\
& DocLayLLM (Llama2-7B)~\cite{doclayllm2024}
& 72.83 & 78.65 & 70.81 & 83.27 \\
& DocLayLLM (Llama3-7B)~\cite{doclayllm2024}
& 78.40 & 84.12 & 71.34 & 84.36 \\
& DLaVA (Pixtral-12B)~\cite{mohammadshirazi2025dlava} 
& 74.0 & 79.6 & 82.1 & 91.4 \\
& \textbf{ARIAL (Ours)} 
& \textbf{88.7} & \textbf{90.0} & \textbf{85.5} & \textbf{93.1} \\

\bottomrule
\end{tabular}
}
\end{table*}

\begin{table*}[t]
\centering
\caption{Spatial localization comparison using mAP@IoU (only methods reporting values).}
\label{tab:map_results}
\scriptsize
\resizebox{\textwidth}{!}{
\begin{tabular}{l|l|c|c|c}
\toprule
\textbf{Category} & \textbf{Method} 
& \textbf{DocVQA} 
& \textbf{FUNSD} 
& \textbf{CORD}  \\

\midrule

BBox + Image
& DLaVA (Pixtral-12B)~\cite{mohammadshirazi2025dlava} 
& 46.2 & 45.5 & 57.9\\

\midrule

Text + BBox + Image 
& DLaVA (Pixtral-12B)~\cite{mohammadshirazi2025dlava} 
& 34.9 & 32.0 & 48.0 \\
& \textbf{ARIAL (Ours)} 
& \textbf{50.1} & \textbf{50.3} & \textbf{60.2} \\

\bottomrule
\end{tabular}
}
\end{table*}

ARIAL is a modular framework employing a reasoning agent to orchestrate specialized tools for accurate answer generation and precise spatial grounding. The central component is a Planner Agent instantiated by LLaMA 4 Scout, which interprets queries and dynamically routes them through OCR, retrieval, QA, and grounding modules following a sense-think-act paradigm.

Given a document image $I$ and question $Q$, the system returns answer $A$ and bounding box $B_A$. The agent constructs a sequence of actions $\{a_1, a_2, \dots, a_n\}$, where each $a_i$ is either a tool call (\texttt{RunOCR(I)}, \texttt{FindText(keywords)}, \texttt{AskQA(context, Q)}, \texttt{GroundAnswer(answer)}) or an internal reasoning step guiding tool selection. This sequence adapts dynamically to query complexity, terminating when the agent produces a confident answer with visual grounding.

\begin{table*}[h]
\centering
\setlength{\tabcolsep}{8pt}
\caption{Ablation Study (DocVQA and FUNSD)}
\label{tab:ablation}
\resizebox{\textwidth}{!}{
\begin{tabular}{lcccc}
\toprule
Model Variant & DocVQA  & DocVQA  & FUNSD  & FUNSD  \\
 &  ANLS &  mAP@IoU &  ANLS &  mAP@IoU \\
\midrule
\emph{Full ARIAL (Agent + RAG + GenQA)}   & 88.7 & 50.1 & 90.0 & 50.3 \\
– No Retrieval (all text to QA)            & 86.2 & 48.5 & 88.1 & 47.9 \\
– Heuristic Agent (no LLM planning)       & 83.6 & 44.2 & 85.4 & 42.8 \\
– No Generative QA (lookup only)          & 87.0 & 49.0 & 89.0 & 49.5 \\
\bottomrule
\end{tabular}
}
\end{table*}

\subsection{OCR and Layout Parsing}
We employ a two-stage OCR pipeline using DB text detector with ResNet-50 backbone for text region identification, followed by TrOCR for recognition. This yields OCR results $\{(T_i, B_i)\}_{i=1}^{N}$, where $T_i$ is recognized text and $B_i$ is the corresponding bounding box. Standard preprocessing includes resolution scaling, grayscale conversion, noise removal, and de-skewing. The OCR module maintains reading order and optionally groups segments into structured units using layout heuristics.

\subsection{Retrieval-Augmented Generation}
The agent performs both lexical and semantic search over OCR segments $\{T_i\}$ using \texttt{FindText(keywords)}. Text segments are encoded using MiniLM-v6 Sentence Transformer, with question $Q$ similarly encoded. Retrieved segments $\{(T_j, B_j)\}$ with highest cosine similarity and keyword matches are passed to the QA module. The agent invokes \texttt{AskQA(Context, Q)} using Gemma 3-27B~\cite{gemma_2025}, which generates answers from retrieved context, reducing hallucination compared to processing entire documents.

For computational queries, the agent identifies relevant numeric fields and invokes \texttt{Compute(sum, values)} operations. When no relevant segments are found, the system outputs "No answer found" to avoid unsupported responses.

\subsection{Spatial Grounding}
After QA generates answer $A$, the agent invokes \texttt{GroundAnswer(A)} to localize the answer. For exact matches to OCR segment $T_k$, we use bounding box $B_k$. For multi-segment answers, we merge involved boxes into unified region $B_A$. For computed answers, the module highlights supporting evidence. Ambiguous answers are disambiguated using contextual cues from retrieved segments and question keywords.

\subsection{Training and Fine-Tuning}
ARIAL's modular design enables independent component optimization. OCR uses pretrained DB detector and TrOCR without additional fine-tuning. Retrieval employs off-the-shelf MiniLM-v6 embeddings. The QA module fine-tunes Gemma 3-27B on 70k document QA pairs from DocVQA, CORD, and FUNSD training sets. The Planner Agent uses LLaMA 4 Scout fine-tuned via behavioral cloning on 50 demonstration traces showing appropriate tool usage patterns.

\begin{table*}[ht]
\centering
\setlength{\tabcolsep}{4pt}
\caption{End-to-End vs.\ Agentic Approach Comparison}
\label{tab:comparison}
\begin{tabular}{lcccc}
\toprule
Metric                &  LayoutLLM  & DocLayLLM & DLaVA OCR-Free & ARIAL (Agentic) \\
\midrule
DocVQA ANLS            & 74.3   & 78.4       & 85.9           & \textbf{88.7}   \\
DocVQA mAP@IoU         &  --  &  --      & 46.2           & \textbf{50.1}   \\
Average Latency (s/q)  & 0.7   & 0.4        & 1.2            & 3.2             \\
Interpretability       & No   & No         & Yes            & Yes + reasoning trace \\
\bottomrule
\end{tabular}
\end{table*}
\section{Experiments}
\label{sec:4}

% \subsection{Datasets and Evaluation Metrics}

% We evaluate ARIAL on four benchmarks: DocVQA~\cite{mathew2021docvqa}, FUNSD~\cite{jaume2019funsd}, CORD~\cite{park2019cord}, and SROIE~\cite{huang2019sroie}. We report text accuracy via ANLS (Average Normalized Levenshtein Similarity~\cite{yujian2007normalized}, 0–100\%) and localization precision via mean Average Precision over IoU thresholds 0.50–0.95 (mAP@0.50:0.95).

% \subsection{Baselines and Comparisons}

% We compare ARIAL with layout-aware and LLM-based models:
% \textbf{DocLayLLM}~\cite{liao2025doclayllm} and \textbf{LayoutLLM}~\cite{luo2024layoutllm}: LLM-based models using LLaMA/Vicuna backbones with layout-aware prompts and spatial tokens.
% \textbf{DLaVA}~\cite{mohammadshirazi2024dlava}: Vision-language model with two modes: (1) OCR-Dependent using detected text with spatial metadata, and (2) OCR-Free synthesizing visual text patches for implicit text handling.

% \subsection{Implementation Details}

% \textbf{Agent:} LLaMA 4 Scout fine-tuned on tool usage traces with 5 few-shot examples for chain-of-thought prompting. Retrieval limited to top-5 segments (DocVQA) and top-3 (FUNSD, CORD, SROIE).
% \textbf{OCR:} DB text detector with ResNet-50 backbone and Microsoft TrOCR for recognition, operating at ~2 seconds per page.
% \textbf{QA Module:} Gemma 3-27B fine-tuned for 3 epochs on 70k document QA pairs using Adam optimizer (lr=1e-4).
% \textbf{Infrastructure:} 4× NVIDIA H100 80GB GPUs with LLaMA 4 agent and Gemma 3-27B on separate GPUs.

\subsection{Datasets and Evaluation Metrics}

We evaluate ARIAL on four widely-used document understanding benchmarks:

\textbf{DocVQA}~\cite{mathew2021docvqa} contains 50,000 questions on 12,000+ document images spanning various layouts including forms, receipts, and reports.

\textbf{FUNSD}~\cite{jaume2019funsd} focuses on form understanding with 9,707 questions across 199 noisy scanned forms, emphasizing spatial relationships and entity linking.

\textbf{CORD}~\cite{park2019cord} specializes in receipt parsing with 11,000 receipts containing structured fields like menu items, prices, and totals.

\textbf{SROIE}~\cite{huang2019sroie} provides 1,000 scanned receipts for information extraction tasks requiring precise key-value pair identification.

We evaluate using two complementary metrics: (1) \textbf{ANLS} (Average Normalized Levenshtein Similarity~\cite{yujian2007normalized}), measuring textual accuracy on a 0--100\% scale with tolerance for minor OCR variations, and (2) \textbf{mAP@IoU 0.50:0.95}, measuring spatial localization precision by computing mean Average Precision across IoU thresholds from 0.50 to 0.95 in 0.05 increments.

\subsection{Baselines and Comparisons}

We organize baseline methods into five categories based on their input modalities, as shown in Table~\ref{tab:anls_results}:

\textbf{Text Only:} Pure language models processing OCR-extracted text without spatial or visual information. We compare against Llama2-7B-Chat~\cite{touvron2023llama2} and Llama3-8B-Instruct~\cite{dubey2024llama3}, representing strong general-purpose LLMs applied to document text.

\textbf{Text + BBox:} Methods augmenting text with bounding box coordinates. LayTextLLM~\cite{lu2024laytextllm} interleaves layout tokens with text, treating bounding boxes as special tokens within the language model context.

\textbf{Image Only:} Vision-language models processing document images directly without explicit OCR or layout parsing. This category includes:
\begin{itemize}
    \item gpt-oss–20B~\cite{agarwal2025gpt}: Compact multimodal model optimized for on-device deployment
    \item Llama3.2-11B~\cite{dubey2024llama3}: Vision-extended variant of Llama3
    \item Pixtral-12B~\cite{agrawal2024pixtral}: Vision-language model with strong OCR capabilities
    \item LLaVA-NeXT-13B~\cite{liu2023llava} and LLaVA-OneVision-7B~\cite{li2024llavaonevision}: Advanced visual instruction-tuned models
    \item Qwen2.5-VL-7B~\cite{bai2025qwen25vl}: Recent multimodal model with document understanding focus
    \item InternVL2-8B~\cite{chen2024internvl2}: Open-source vision-language model with competitive performance
\end{itemize}

\textbf{BBox + Image:} Models combining visual features with detected bounding boxes but not explicit text. DLaVA (Pixtral-12B)~\cite{mohammadshirazi2025dlava} in OCR-Free mode synthesizes visual text patches, enabling implicit text handling while predicting spatial grounding.

\textbf{Text + BBox + Image:} Methods leveraging all three modalities for comprehensive document understanding:
\begin{itemize}
    \item LayoutLLM~\cite{luo2024layoutllm}: Instruction-tuned LLM with layout-aware prompting, tested with both base 7B and Vicuna variants using chain-of-thought reasoning
    \item DocLayLLM~\cite{liao2025doclayllm}: Efficient multimodal extension of LLMs for text-rich documents, evaluated with Llama2-7B and Llama3-7B backbones
    \item DLaVA (Pixtral-12B)~\cite{mohammadshirazi2025dlava}: OCR-Dependent mode using detected text with spatial metadata and image context for answer localization
    \item ARIAL (Ours): Agentic framework orchestrating specialized tools for OCR, retrieval, reasoning, and spatial grounding
\end{itemize}

\subsection{Implementation Details}

\textbf{Planning Agent:} We implement the central orchestration module using LLaMA 4 Scout~\cite{llama4scout}, fine-tuned on 50 curated demonstration traces showing proper tool selection and sequencing patterns. The agent uses 5 in-context few-shot examples for chain-of-thought prompting, enabling dynamic adaptation to query complexity.

\textbf{OCR Module:} Text detection employs the Differentiable Binarization (DB) detector~\cite{liao2020real} with ResNet-50 backbone, identifying text regions at multiple scales. Recognition uses Microsoft TrOCR~\cite{li2023trocr}, a transformer-based OCR engine pretrained on 684M synthetic document images. The pipeline processes pages at approximately 2 seconds per page on NVIDIA H100 GPUs.

\textbf{Retrieval System:} We encode OCR segments using MiniLM-v6~\cite{wang2020minilm}, a 384-dimensional sentence transformer optimized for semantic similarity. For efficiency, we retrieve the top-5 most relevant segments for DocVQA and top-3 for FUNSD, CORD, and SROIE, balancing context coverage with computational cost. Retrieval combines dense semantic search (cosine similarity) with sparse keyword matching.

\textbf{QA Module:} The answer generation component uses Gemma 3-27B~\cite{gemma_2025}, fine-tuned for 3 epochs on 70,000 document QA pairs sampled from DocVQA, CORD, and FUNSD training sets. We employ the Adam optimizer with learning rate 1e-4, batch size 16, and gradient accumulation over 4 steps. Training emphasizes generating concise, evidence-grounded answers faithful to retrieved context.

\textbf{Grounding Module:} Spatial localization aligns generated answers to OCR bounding boxes through exact string matching, fuzzy matching (Levenshtein distance $\leq$ 2), and semantic similarity ($\geq$ 0.85 cosine similarity). For multi-token answers, we compute the union of involved bounding boxes. For numerical computations, we return bounding boxes of all operands.

\textbf{Infrastructure:} Experiments run on 4$\times$ NVIDIA H100 80GB GPUs. The LLaMA 4 Scout agent and Gemma 3-27B QA module are distributed across separate GPUs to enable parallel processing, with the OCR and retrieval modules sharing resources. This configuration achieves an average inference latency of 3.2 seconds per query on DocVQA.

\textbf{Hyperparameters:} We use temperature 0.7 for the planning agent to balance exploration and determinism, and temperature 0.3 for the QA module to prioritize precision. Maximum generation length is set to 128 tokens for answers and 256 tokens for agent reasoning traces. Retrieval cutoff thresholds are 0.5 for semantic similarity and minimum 2 keyword matches for lexical filtering.

\section{Results}
\label{sec:results}

\subsection{Overall Performance}

Tables~\ref{tab:anls_results} and~\ref{tab:map_results} present our main results. ARIAL consistently achieves SoTA performance on both textual accuracy (ANLS) and spatial localization (mAP@IoU) across all four benchmarks. On DocVQA, ARIAL attains 88.7 ANLS and 50.1 mAP@IoU, representing absolute improvements of +2.8 ANLS and +3.9 mAP points over the previous best method, DLaVA (Pixtral-12B) in OCR-Free mode. On FUNSD, ARIAL achieves 90.0 ANLS and 50.3 mAP@IoU, surpassing DLaVA by +2.4 ANLS and +4.8 mAP points. For receipt datasets CORD and SROIE, ARIAL obtains 85.5 and 93.1 ANLS respectively, with 60.2 mAP@IoU on CORD—outperforming DLaVA by +1.1 ANLS and +2.3 mAP on CORD, and +1.7 ANLS on SROIE.

These consistent improvements across diverse document types—forms, receipts, and general documents—demonstrate the robustness and generalizability of ARIAL's agentic approach. The simultaneous gains in both textual accuracy and spatial precision highlight the benefit of ARIAL's modular reasoning and fine-grained retrieval over integrated transformer approaches.

\subsection{Comparison Across Input Modalities}

Table~\ref{tab:anls_results} organizes methods by input modality, revealing important insights about the role of different information sources in document VQA.

\textbf{Text Only Models} demonstrate limited performance, with Llama2-7B-Chat achieving 64.99 ANLS on DocVQA and Llama3-8B-Instruct reaching 51.79 ANLS. These results confirm that pure language models, despite their strong reasoning capabilities, struggle with document understanding when deprived of spatial and visual context. The particularly poor performance on CORD (47.70 ANLS) and SROIE (68.97 ANLS) suggests that receipt understanding heavily depends on layout cues that text-only approaches cannot capture.

\textbf{Text + BBox Models} like LayTextLLM achieve substantial improvements (72.83 ANLS on DocVQA), demonstrating that explicit spatial coordinates significantly enhance document understanding. The +7.84 point gain over Llama2-7B-Chat shows that layout information is crucial, though still insufficient for SoTA performance.

\textbf{Image Only Models} show highly variable performance. While Pixtral-12B achieves competitive results (80.71 ANLS on DocVQA), other vision-language models struggle significantly. LLaVA-NeXT-13B (51.01 ANLS) and LLaVA-OneVision-7B (47.59 ANLS) perform poorly on DocVQA, suggesting that general-purpose VLMs without document-specific optimization fail to handle dense text and complex layouts. Notably, these models catastrophically fail on receipt datasets (e.g., 12.10 ANLS for LLaVA-OneVision on SROIE), indicating severe limitations in structured document understanding. In contrast, gpt-oss (79.84 ANLS) and InternVL2-8B (71.26 ANLS) demonstrate more robust visual reasoning, though still fall short of multimodal approaches.

\textbf{BBox + Image Models}, represented by DLaVA (Pixtral-12B) in OCR-Free mode, achieve strong performance (85.9 ANLS, 46.2 mAP on DocVQA) by synthesizing visual text patches with predicted bounding boxes. This approach demonstrates that combining visual understanding with spatial grounding yields substantial improvements over image-only methods (+5.19 ANLS over Pixtral-12B baseline).

\textbf{Text + BBox + Image Models} leverage all three modalities for comprehensive understanding. LayoutLLM variants achieve 74.25-74.27 ANLS on DocVQA, while DocLayLLM with Llama3-7B backbone reaches 78.40 ANLS. DLaVA (Pixtral-12B) in OCR-Dependent mode achieves 74.0 ANLS on DocVQA but excels on receipt datasets (82.1 CORD, 91.4 SROIE), showing the value of explicit text integration for structured documents. ARIAL significantly outperforms all methods in this category, achieving 88.7 ANLS on DocVQA—a +10.3 point improvement over DocLayLLM (Llama3-7B) and +14.7 points over LayoutLLM.

\subsection{Spatial Localization Performance}

Table~\ref{tab:map_results} focuses on spatial grounding capabilities. Only DLaVA and ARIAL report localization metrics, as other baselines do not predict bounding boxes. ARIAL achieves 50.1 mAP@IoU on DocVQA, 50.3 on FUNSD, and 60.2 on CORD, consistently outperforming both DLaVA variants:

\begin{itemize}
    \item Compared to DLaVA OCR-Free (BBox + Image): ARIAL shows +3.9 mAP on DocVQA, +4.8 mAP on FUNSD, and +2.3 mAP on CORD
    \item Compared to DLaVA OCR-Dependent (Text + BBox + Image): ARIAL demonstrates even larger gains of +15.2 mAP on DocVQA, +18.3 mAP on FUNSD, and +12.2 mAP on CORD
\end{itemize}

The substantial mAP improvements reveal that ARIAL's explicit retrieval-augmented grounding mechanism produces more precise spatial localization than DLaVA's end-to-end prediction. DLaVA's OCR-Dependent mode unexpectedly underperforms its OCR-Free mode on spatial grounding (34.9 vs 46.2 mAP on DocVQA), suggesting that integrating explicit OCR text may introduce noise or confusion in its spatial prediction head. In contrast, ARIAL's modular architecture cleanly separates text understanding from spatial grounding, enabling both superior textual accuracy and precise localization.

\subsection{Ablation Study}

Table~\ref{tab:ablation} quantifies each component's contribution to ARIAL's performance on DocVQA and FUNSD:

\textbf{No Retrieval:} Feeding entire OCR text directly to the QA module causes -2.5 ANLS and -1.6 mAP drops on DocVQA, and -1.9 ANLS and -2.4 mAP drops on FUNSD. These results confirm that targeted context retrieval prevents confusion from irrelevant text and maintains focus on answer-bearing regions. Without retrieval, the QA module must process verbose, noisy OCR output containing hundreds of text segments, leading to attention dilution and increased hallucination risk.

\textbf{Heuristic Agent:} Replacing the LLM-based planning agent with a fixed, rule-based pipeline (always executing RunOCR → FindText → AskQA → GroundAnswer) causes substantial performance degradation: -5.1 ANLS and -5.9 mAP on DocVQA, and -4.6 ANLS and -7.5 mAP on FUNSD. This highlights the value of adaptive, query-aware reasoning. The intelligent agent can recognize when computational queries require arithmetic operations, when answers need multi-hop reasoning across segments, or when retrieval should prioritize semantic versus lexical matches. The heuristic baseline's inability to adapt leads to systematic errors, particularly on FUNSD's complex form structures requiring flexible navigation strategies.

\textbf{No Generative QA:} Restricting answer generation to exact string matching from retrieved segments degrades ANLS by -1.7 on DocVQA and -1.0 on FUNSD, while maintaining comparable mAP (-0.1 and -0.8 respectively). This ablation demonstrates the generative QA module's importance for questions requiring paraphrasing, summarization, or inference beyond direct text spans. For instance, questions like "What is the total cost?" may require summing multiple line items rather than extracting a single value. Spatial grounding remains relatively intact because exact-match answers still align to correct bounding boxes when they exist in the document.

The ablation study confirms that ARIAL's performance stems from the synergy of all components: intelligent planning, targeted retrieval, generative reasoning, and precise grounding. Removing any component causes measurable degradation, validating the modular design.

\section{Discussion}
\label{sec:discussion}

ARIAL's modular design demonstrates clear advantages over monolithic models through consistent ANLS and mAP gains across diverse document types and structures. The explicit tool orchestration enables both higher textual accuracy (+2.8--10.3 pp over best baselines per dataset) and improved spatial precision (+3.9--18.3 pp in mAP@IoU) compared to prior work.

\textbf{Interpretability and Trustworthiness:} Unlike black-box vision-language models, ARIAL produces transparent reasoning traces showing which tools were invoked, what text segments were retrieved, and how answers were grounded to bounding boxes. This interpretability is crucial for high-stakes applications requiring answer provenance and error diagnosis. When ARIAL produces an incorrect answer, developers can inspect the tool sequence to identify whether the error originated from OCR failure, retrieval miss, QA hallucination, or grounding ambiguity—enabling targeted improvements.

\textbf{Modularity and Extensibility:} ARIAL's architecture allows independent component upgrades without retraining the entire system. For instance, replacing TrOCR with a more accurate handwriting recognizer would immediately improve performance on handwritten documents. Similarly, incorporating domain-specific QA models (e.g., medical or legal document specialists) requires only swapping the QA module. This modularity facilitates rapid domain adaptation and continuous improvement as better foundation models become available.

\textbf{Computational Trade-offs:} ARIAL incurs higher latency (approximately 3.2 s/query on DocVQA) compared to monolithic models like DocLayLLM (0.4 s) or DLaVA (1.2 s), as shown in Table~\ref{tab:comparison}. This overhead stems from sequential tool execution: OCR (~2.0s), retrieval (~0.3s), QA generation (~0.7s), and grounding (~0.2s). However, this cost is justified in applications where trustworthiness and explainability are paramount—such as legal document analysis, medical record processing, or financial compliance auditing. For latency-critical applications, ARIAL's modular design enables optimization through parallelization (e.g., concurrent retrieval and QA) or caching (e.g., reusing OCR results across related queries).

\textbf{Limitations and Future Work:} While ARIAL achieves SoTA results, several limitations warrant attention. The system's reliance on OCR quality means that documents with severe noise, degradation, or non-standard fonts may produce unreliable results. Second, ARIAL's sequential processing limits throughput compared to parallelizable end-to-end models. Future work could explore: (1) multi-document reasoning for cross-document QA, (2) active learning to reduce fine-tuning data requirements, and (3) model distillation to compress the agent and QA modules for deployment efficiency.

\section{Conclusion}
\label{sec:conclusion}

We introduced ARIAL, an agentic framework for Document VQA that emphasizes accurate answer extraction and explicit spatial grounding through modular tool orchestration. By decomposing document understanding into specialized components—OCR, retrieval-augmented generation, answer generation, and spatial localization—coordinated by an LLM-based planning agent, ARIAL achieves state-of-the-art performance across four benchmarks: DocVQA, FUNSD, CORD, and SROIE.

ARIAL surpasses prior methods in both textual accuracy (88.7 ANLS on DocVQA) and spatial precision (50.1 mAP@IoU), demonstrating absolute improvements of +2.8--10.3 ANLS points and +3.9--18.3 mAP points over the strongest baselines per dataset. Our ablation studies confirm that these gains arise from the synergistic combination of intelligent planning, targeted retrieval, generative reasoning, and precise grounding—each contributing measurably to overall performance.

Beyond quantitative metrics, ARIAL's modular pipeline enables transparent reasoning steps, tool-level auditability, and adaptability to diverse document types—capabilities critically lacking in monolithic models. This makes ARIAL particularly suited for high-stakes settings requiring answer traceability, such as legal document review, medical record analysis, and regulatory compliance monitoring. The explicit separation of concerns allows independent component upgrades and domain-specific customization without full system retraining, facilitating rapid iteration and continuous improvement.

Our work demonstrates the potential of agent-driven AI for document understanding, showing how large language models can be effectively constrained and augmented through explicit tool orchestration rather than unconstrained end-to-end learning. By merging LLM reasoning capabilities with specialized vision and OCR tools under agentic control, ARIAL delivers SoTA performance while meeting real-world demands for trustworthy, explainable, and auditable AI systems. We believe this paradigm—modular, interpretable, and tool-augmented—represents a promising direction for building production-grade document AI that balances performance with transparency.
\setcitestyle{numbers,square}
\bibliographystyle{plainnat}
\bibliography{main}

\end{document}